\documentclass[letterpaper, 10 pt, conference]{ieeeconf}
\IEEEoverridecommandlockouts                              

\usepackage{cite}
\usepackage{amsmath,amssymb,amsfonts}
\usepackage{algorithmic}
\usepackage{graphicx}
\usepackage{textcomp}
\usepackage{xcolor}
\usepackage{svg}
\usepackage{bm}
\usepackage{stfloats} 
\usepackage[normalem]{ulem}

\usepackage{hyperref}
\def\BibTeX{{\rm B\kern-.05em{\sc i\kern-.025em b}\kern-.08em
    T\kern-.1667em\lower.7ex\hbox{E}\kern-.125emX}}

\newcommand{\R}{\mathbb{R}}
\newcommand{\bd}[1]{\mathbf{#1}}

\usepackage{listings}
\usepackage{xcolor} 

\begin{document}

\title{CRISP - \underline{C}ompliant \underline{R}OS2 Controllers for Learn\underline{i}ng-Ba\underline{s}ed Manipulation \underline{P}olicies and Teleoperation
\thanks{
The authors are with the Technical University of Munich (TUM), Germany; the TUM School of Computation, Information and Technology, Department
of Computer Engineering, Learning Systems and Robotics Lab and the TUM Munich
Institute of Robotics and Machine Intelligence (MIRMI).
Email: \{daniel.sanjose.pro; oliver.hausdoerfer; maximilian.doesch; martin.schuck; angela.schoellig\}@tum.de.}
}

\author{Daniel San José Pro,
Oliver Hausd\"orfer,
Ralf R\"omer,
Maximilian D\"osch,
Martin Schuck
and Angela P. Schoellig
}

\maketitle

\begin{figure*}[!t]
    \centering
    \includegraphics[width=1\textwidth]{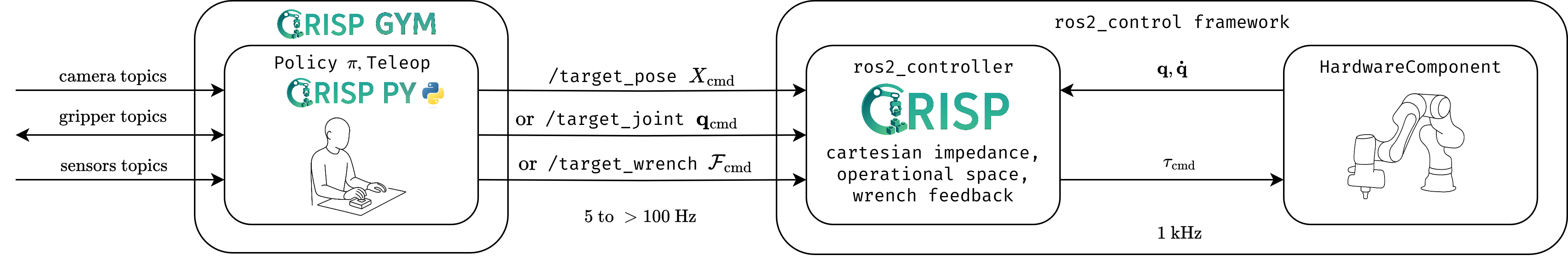}
    \caption{\href{https://github.com/utiasDSL/crisp_controllers}{CRISP} architecture overview and integration with robots using ROS2. We provide \href{https://github.com/utiasDSL/crisp_py}{CRISP\_PY}, an easy-to-use Python interface to the CRISP controllers, as well as Gymnasium~\cite{towers2024gymnasium} environments we use for data collection and policy deployment in \href{https://github.com/utiasDSL/crisp_gym}{CRISP\_GYM}. The user can publish any of the available target commands at arbitrary frequencies via ROS2 topics, and the robot will track the most recent target. High-level learning-based policies such as vision-language-action (VLA) models typically use the \texttt{target\_pose} topic. We run the CRISP controllers on a real-time enabled Linux workstation at 1 kHz, while  CRISP\_GYM with the VLA model runs on a separate workstation in the same network, communicating via ROS2 topics. CRISP\_GYM further easily integrates robotic grippers, cameras, and additional sensors required for the task via ROS2 topics.}
    \label{fig:overview}
\end{figure*}

\begin{abstract}
Learning-based controllers, such as diffusion policies and vision-language action models, often generate low-frequency or discontinuous robot state changes. Achieving smooth reference tracking requires a low-level controller that converts high-level targets commands into joint torques, enabling compliant behavior during contact interactions. We present \textsc{CRISP}, a lightweight C++ implementation of compliant Cartesian and joint-space controllers for the ROS2 control standard, designed for seamless integration with high-level learning-based policies as well as teleoperation. The controllers are compatible with any manipulator that exposes a joint-torque interface. Through our Python and Gymnasium interfaces, \textsc{CRISP} provides a unified pipeline for recording data from hardware and simulation and deploying high-level learning-based policies seamlessly, facilitating rapid experimentation. The system has been validated on hardware with the Franka Robotics FR3 and in simulation with the Kuka IIWA14 and Kinova Gen3. Designed for rapid integration, flexible deployment, and real-time performance, our implementation provides a unified pipeline for data collection and policy execution, lowering the barrier to applying learning-based methods on ROS2-compatible manipulators. Detailed documentation is available at the \href{https://utiasDSL.github.io/crisp_controllers}{project website}.

\end{abstract}


\section{Introduction}
In recent years, learning-based policies for robotic manipulation have become increasingly prevalent.
These approaches commonly command changes in the robot state as small trajectories over a short horizon, also called action chunks, or low-frequent targets, often at around 5 to 10 Hz, without requiring a traditional perception and path planning stack~\cite{brohan2023rt2, zhao2023finemulti, chi2024diffusionpolicy, black2024pi0visionlanguageactionflowmodel, kim2025openvla}.
Such paradigms require low-level controllers that can respond to command changes during deployment rather than following precomputed smooth trajectories, which has been the typical mode of operation in manipulation in the past \cite{Brady-1982-15599, lynch2017modern}. Moreover, many challenging manipulation tasks often involve interaction with the environment, including contact-rich behaviors.

Learning-based applications demand responsive, low-latency controllers capable of reacting to both user or high-level policy input and to a dynamically changing environment. 
While ROS and ROS2 have generally simplified robotics software development, current ROS2 tooling for such tasks often introduces significant overhead. 
At the same time, several robot manufacturers, including Franka Robotics, Kinova, Duatic, and ROBOTIS, together with the broader ROS2 community, have contributed high-quality \texttt{ros2\_control}~\cite{ros2_control} hardware interfaces that are actively maintained and support direct torque control for a wide range of manipulators. 
This convergence creates an opportunity to implement modular, low-level controllers that integrate directly with the ROS2 control interface and are well suited to the requirements of high-level learning-based policies.

\textbf{Contributions.}  
We present \textsc{CRISP}, a C++ implementation with Python interfaces of compliant Cartesian and joint-space controllers for robotic manipulators compatible with the ROS2 and \texttt{ros2\_control} frameworks.  
Our main contributions are:  
\begin{itemize}
    \item \textbf{Robot-agnostic, compliant Cartesian and joint-space controllers} that leverage ROS2 standard interfaces to ensure broad compatibility.
    \item \textbf{A unified pipeline for data collection and deployment of learning-based policies} through our Python interface \textsc{CRISP\_PY} and Gymnasium interface \textsc{CRISP\_GYM}.
\end{itemize}
\section{Related Work}

Traditionally, executing complex robot manipulation tasks relies on a path planning and perception pipeline.
Established frameworks such as MoveIt~\cite{coleman2014reducingbarrierentrycomplex} provide comprehensive planning capabilities; however, they often depend on extensive software stacks that require users to have in-depth knowledge for effective integration into their systems.
In contrast, recent learning-based approaches focus on producing small chunks or low-frequency target poses rather than full trajectories, which reduces system complexity and necessitates a fundamentally different architectural framework.

Several existing efforts address discontinuous Cartesian control within ROS. Table~\ref{tab:comparison} compares our implementation with other available options.
The Cartesian controllers developed by the FZI Research Center for Information Technology (FZI~CC)~\cite{FDCC} support position and velocity control but do not provide torque-based control. Other controllers, such as the Cartesian Impedance Controller (CIC)~\cite{mayr2024cartesian}, \texttt{compliant\_controllers}~\cite{mitchell2025taskjointspacedualarm}, or \texttt{serl\_franka\_controllers}~\cite{serl_franka_controllers}, implement Cartesian impedance control but are only available for ROS1 or are specific to a single manipulator type.
Additional user-developed tools have also emerged~\cite{zhu2022viola, elsner2023taming, schneider_franky}, but these are typically robot specific or lack \texttt{ros2\_control} integration.
The emerging LeRobot framework~\cite{cadene2024lerobot} improves accessibility for learning-based manipulation, but does not yet support a wide range of manipulators and lacks full integration with ROS2.

In contrast to existing frameworks, we present robot-agnostic controllers that integrate seamlessly with the ROS2 standard and support both data collection and deployment of learning-based policies.


\begin{table}
    \centering
    \begin{tabular}{lccc}
    \hline
            & FZI CC & CIC & Ours \\
    \hline
      ROS1  & x      &  x  & \\
      ROS2  &  x     &    & x \\
      Torque/effort interface &  & x  & x \\
      Robot agnostic & x  & x & x \\
      Highly Configurable & (x) & x & x \\
      Cartesian Impedance &  & x & x \\
      Nullspace Biasing &  & x & x \\
      Force control &  & x & x \\
      Joint limits barrier &  &  & x \\
      Friction Compensation &  &  & x \\
      Operation Space Control &  &  & x \\
      Built with \texttt{pinocchio}\footnotemark &  &  & x \\
      Gymnasium Wrapper & & & x \\ 
    \hline\\
    \end{tabular}
    
    \caption{Comparison to the closest available controller implementations}
    \label{tab:comparison}
    \vspace{-14pt}
\end{table}

\section{Implementation}
\footnotetext{ROS native, well-tested kinematics and rigid-body dynamics library based on Lie groups.}

To ensure robot agnosticism, we implement compliant controllers in C++ for the \texttt{ros2\_control} interface, capable of running at 1~kHz on a joint-torque interface. These controllers are designed to be responsive and robust to the discontinuous command updates typically generated by high-level learning-based policies. As shown in the overview in ~\autoref{fig:overview}, users can publish target commands at any frequency using \textsc{CRISP\_PY} and \textsc{CRISP\_GYM}, and the controller will continuously track the most recent command.

The controllers implement joint limit barriers and include friction compensation for smooth teleoperation. All required rigid-body computations, as well as parsing of the robot structure from the standard Universal Robot Description File (URDF), are handled by \href{https://github.com/stack-of-tasks/pinocchio}{\texttt{pinocchio}}~\cite{carpentier2019pinocchio}. The \texttt{pinocchio} library efficiently implements rigid-body dynamics, is distributed with ROS2 binaries, and supports modern automatic differentiation in addition to other features required for advanced control algorithms.

To enable configurability and ease of integration into other software stacks, all controller parameters are dynamically configurable via standard ROS2 parameter services. Examples of configurable parameters include stiffness, damping, nullspace stiffness, torque rate limits, and error limits. Several default configurations are provided in our repositories.

\subsection{Task Torque}

The task torque $\bm \tau_\text{task}$ refers to the desired torques to be applied to solve the task of following a target pose $(\bd x_\text{target},\bd R_\text{target})\in\R^3\times SO(3)$ which is represented with respect to the base frame. 
It attempts to minimize the error $\bd e =(\bd e_\text{pos}, \bd e_\text{rot})\in\R^6$ with respect to the current pose $(\bd x_\text{current}(\bd q), \bd R_\text{current}(\bd q))$ computed using forward kinematics.
The error can computed with respect to the base as
\begin{align}
    \bd{e}_\text{pos}^\text{} &= \bd{x}_\text{target} - \bd{x}_\text{current}(\bd q) \\
    \bd{e}_\text{rot}^\text{}&= \text{Log}(\bd{R}_\text{target}\bd{R}_\text{current}^{\top}(\bd q))
\end{align}
or with respect to the end effector-frame as
\begin{align}
    \bd{e}_\text{pos}^\text{}&= \bd R_\text{current}^\top(\bd x_\text{target} - \bd x_\text{current}(\bd q)), \\
    \bd{e}_\text{rot}^\text{}& = \text{Log}(\bd R_\text{current}^\top(\bd q) \bd R_\text{target}),
\end{align}
where Log is the logarithmic map used to map a rotation difference to the tangent space of $SO(3)$. This can be seen as an angular velocity that turns the end-effector rotation towards the target rotation~\cite{solà2021microlietheorystate}.
Defining the error in different frames allows the user to define anisotropic stiffness and damping behavior of the controller with respect to the end-effector or base frame.

\subsubsection{Cartesian Impedance (CI) Controller}

This controller implements a virtual spring between the end-effector and the reference pose and is well suited for contact-rich, compliant tasks, as it adapts well to contacts \cite{hogan_impedance_1984}:
\begin{equation}
    \bm\tau_\text{task} = \bd J^\top \left(\bd{K}_\text{p}\bd e - \bd{K}_\text{d} \bd J\dot{\bd q}\right),
\end{equation}
where $\bd J$ denotes the geometric Jacobian with respect to the base or end-effector frame computed from the provided URDF and $\bf{K}_\text{p}$, $\bf{K}_\text{d}$ are the stiffness and damping matrices respectively chosen by the user.

\subsubsection{Operational Space Controller (OSC)} Compared to the CI controller, OSC accounts for the dynamics, in particular the inertia, in the control law~\cite{khatib_unified_1987}. The OSC tends to be more precise but less compliant in the presence of contact. The task torque is given by
\begin{equation}
    \bm \tau_\text{task} = \bd J^\top\bm\Lambda(\bd q,\dot{\bd q}) \left(\bd{K}_\text{p} \bd e - \bd{K}_\text{d} \bd J\dot{\bd q}\right),
\end{equation}
where $\bm\Lambda(\bd q,\dot{\bd q})=(\bd J^{\dagger\top}\bd M^{-1}\bd J^{\dagger})^{-1}$ denotes the task-space mass matrix computed using the mass matrix $\bd M$. The mass matrix is determined with the provided URDF using \texttt{pinocchio}.
OSC does not compensate for the Coriolis and gravity terms in the task space. Instead, these are dynamically decoupled and directly compensated in joint-space, since it makes computations easier and the performance is similar to the original OSC \cite{nakanishi_operational_2008}. 

\subsection{Null Space Control}

For redundant manipulators, we provide the option to control in the null space of the task, which is useful to limit unnecessary motions, avoid singularities, and stabilize the joints to the resting position. This can be achieved by projecting a secondary control law $\bm\tau_\text{secondary}$ with a null space projector $\mathbf{N} \in\mathbb{R}^{n\times n}$ to a subspace that does not interfere with the primary task \cite{nullspace2015}. The secondary task in our control law is a joint impedance controller computed as 
\begin{equation}
    \bm\tau_\text{secondary} = \bd{K}_\text{p,s} (\bd q_
    \text{target}-\bd q) +\bd{K}_\text{d,s}(\bd{\dot{q}}_\text{target}-\bd{\dot{q}}).
\end{equation}

We provide different projectors to compute the null space torque $\bm\tau_\text{ns}=\bf{N}\bm\tau_\text{secondary}$: (1) a static null space projector $\bf{N}=\bd I- \bd J^\top (\bd J^\dagger)^\top$, (2) a dynamic null space projector $\bf{N}=\bd I- \bd J^\top \bd{\bar{J}}^\top$, where $\bd{\bar{J}}=\bd M^{-1} \bd J^\top (\bd J \bd M^{-1}\bd J^\top)^{-1}$ is the generalized inverse Jacobian \cite{khatib_unified_1987}, and (3) an identity null space projector $\bf{N}=\bd I$ that can be used to directly control the manipulator with joint commands if the task stiffness  $\bd{K}_\text{p}$ and damping $\bd{K}_\text{d}$  have been set to 0.
The dynamic null space projector leads to better primary task tracking compared to the static null space projector, provided that a sufficiently accurate model (URDF) of the robot is available~\cite{albu-schaffer_cartesian_2003}.

\subsection{Joint Barrier}

In order to avoid reaching the joint limits of the robot and prevent damage, we add joint barriers that apply an increasing torque as the joint gets close to the limits by $\epsilon$:

\begin{equation}
    \tau_{i,\text{joint}} = \begin{cases}
        -{K}_{i,\text{joint}} (q_{i,\text{max}} - q_i) & \text{if } q_i>\theta_{i,\text{max}} - \epsilon, \\
        -{K}_{i,\text{joint}} (q_{i,\text{min}} - q_i) & \text{if } q_i<\theta_{i,\text{min}} + \epsilon, \\
        0 & \text{otherwise.}
        
    \end{cases}
\end{equation}

\subsection{Gravity and Coriolis Compensation}
We efficiently compute the gravity compensation and coriolis term using \texttt{pinocchio}, which uses the Recursive Newton Euler Algorithm (RNEA) to solve the Inverse Dynamics problem $\tau=RNEA(\bd q,\bd {\dot q},\bd {\ddot q},\bm{\mathcal F}_\text{ext})$ to determine $\bm \tau_\text{gravity} = \bd g(\bd q) = RNEA(\bd q, \bd 0, \bd 0, \bd 0)$ and $\bm\tau_\text{coriolis} = \bd C(\bd q,\bd{\dot q}) \bd{\dot q}$.
For the user, it is therefore important to provide the inertial parameters in the URDF file if these are not included by default. In case a manipulator already provides gravity compensation, it can be removed from the control law.

\subsection{Friction Compensation}

We provide a term to compensate for friction effects that are unmodeled terms in the equations of motion for rigid bodies. The friction can be modeled as a combination of static Coulomb friction, viscous friction, and a friction offset. 
We provide a simplified model as used in \cite{FrankaPandaDynParams} to estimate the Franka Emika Robot friction term:
\begin{equation}
\bm \tau_\text{friction} = \bm\varphi_1 \left(\sigma(\bm \varphi_2(\bd{\dot q}+\bm\varphi_3)) - \sigma(\bm\varphi_2\bm\varphi_3)\right)
\end{equation}
with $\sigma(u) = \frac{1}{1+\exp{(-u)}}$.
This model of friction neglects viscous terms and uses a sigmoidal function instead of a sign function to avoid sudden discontinuities around zero velocities.

While we provide friction parameters for the Franka Robotics FR3~\cite{FrankaPandaDynParams}, these parameters must be determined for each robot by the user or the term omitted entirely. In practice, this term is particularly useful for teleoperation, as it makes the operator feel less unnatural resistance when moving the robot. 

\subsection{Target Wrench}

Using our implementations, it is possible to command a wrench at the end-effector to apply external forces and torques. The term is computed as $\bm\tau_\text{wrench} = \bd J^\top\bm{\mathcal{F}}_\text{target}$,
where $\bm{\mathcal{F}}\in\R^6$ denotes the desired wrench. Wrench control is useful for tasks that demand careful interaction with the environment, require specific torques to be applied such as in screw driving~\cite{lynch2017modern}, compensate for the weight of end-effector extensions, or simply drive the robot with a wrench instead of Cartesian poses. 

\subsection{Final Control Law}

Our final control law is a sum of all the previous terms:
\begin{equation}
\begin{split}
\bm\tau_\text{cmd} = & \;\bm\tau_\text{task}+\bm\tau_\text{ns}+\bm\tau_\text{joint} + \bm\tau_\text{gravity} \\
    & + \bm\tau_\text{coriolis}+\bm\tau_\text{friction} +\bm\tau_\text{wrench}
\end{split}
\end{equation}

In our implementation, each term can be switched on and off depending on the task, giving full flexibility to the user.

\subsection{Additional Features}



We provide additional safety features including joint torque and rate limits, exponential filtering of targets to avoid fast movements, and tracking error clipping for precise yet safe control when targets are distant~\cite{serl_franka_controllers}. For teleoperation, we implement force-torque feedback controllers where follower forces are reflected to the leader as $\bm{\tau}_\text{fb} = -k_\text{p,fb}\mathbf{J}^\top \mathcal{F}_\text{follower} - k_{\text{d,fb}} \dot{\bf{q}}_\text{leader}$.

\begin{figure}[t]
    \centering
    \includegraphics[width=1\linewidth]{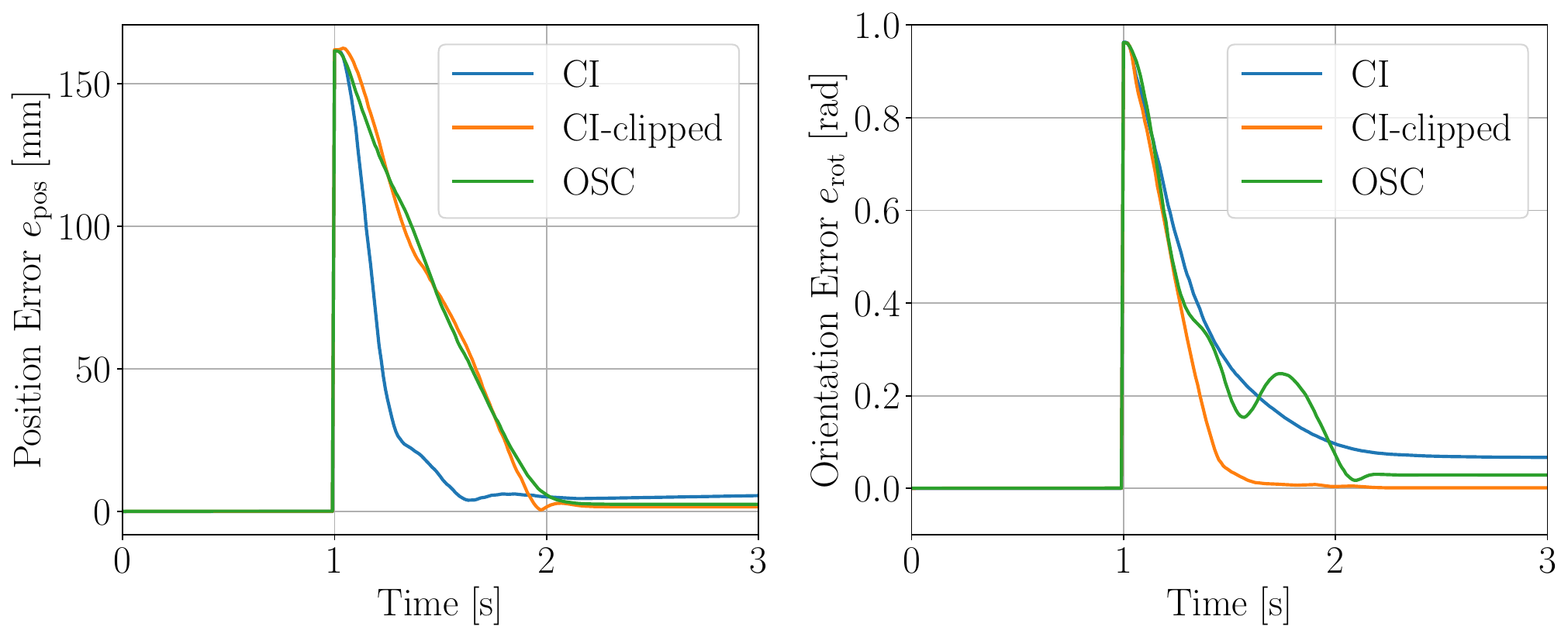}

    \caption{\textbf{Robot-agnostic Cartesian control [I]:}  
    Error evolution for tracking a discontinuous target pose using a Franka Emika FR3 on hardware with \textsc{CRISP}. We set a new target pose at $t=1$~s. For reproducibility, the controller parameters used are available on our website. Note that exact tracking performance depends on controller parameterization, which should be chosen based on the specific task. \textbf{Left:} Position error evolution with steady-state errors of 5.54~mm, 4.73~mm, and 0.81~mm for OSC, CI, and CI-clipped, respectively. \textbf{Right:} Rotational error evolution with steady-state errors of 0.0998~rad, 0.0532~rad, and 0.0029~rad for OSC, CI, and CI-clipped, respectively.
    \vspace{-4pt}
    }
    \label{fig:sporadic_poses}
\end{figure}

\begin{figure}
    \centering
    \includegraphics[width=1\linewidth]{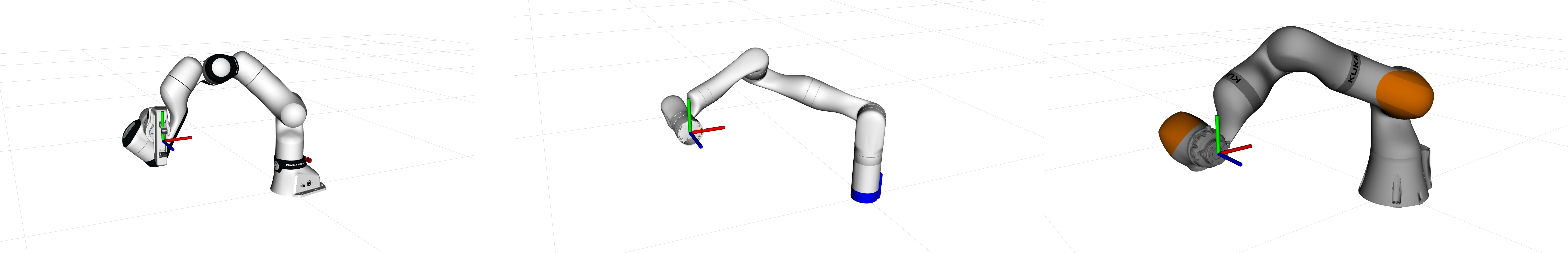}
    \caption{\textbf{Robot agnostic Cartesian control [II]:} We demonstrate our controllers on a Franka FR3 on hardware, and on multiple manipulators in simulation. \textbf{Left:} Franka Robotics FR3. \textbf{Middle:} Kinova Gen3. \textbf{Right}: Kuka IIWA 14.}
    \label{fig:robots}
    \vspace{-12pt}
\end{figure}

\section{Evaluation}

The controllers have been validated on hardware using the Franka Robotics FR3 and in simulation with the Kuka IIWA14 and Kinova Gen3, as shown in Figure~\ref{fig:robots}. Videos of these experiments are available on the \href{https://utiasDSL.github.io/crisp_controllers}{project website}. In this section, we present results for three representative tasks performed with our controllers. All experiments were conducted on a real Franka Robotics FR3. The exact tracking performance depends on the parameters selected by the user for the specific task as well as the manipulator used.

First, we evaluate the tracking capabilities by issuing a randomly generated target pose within a safe motion range and monitoring the evolution of the tracking error over time, as shown in~\autoref{fig:sporadic_poses}. This experiment illustrates the tracking performance for a fixed target pose under different parameter configurations available to the user.

\begin{figure}[t]
    \centering
    \includegraphics[width=1\linewidth]{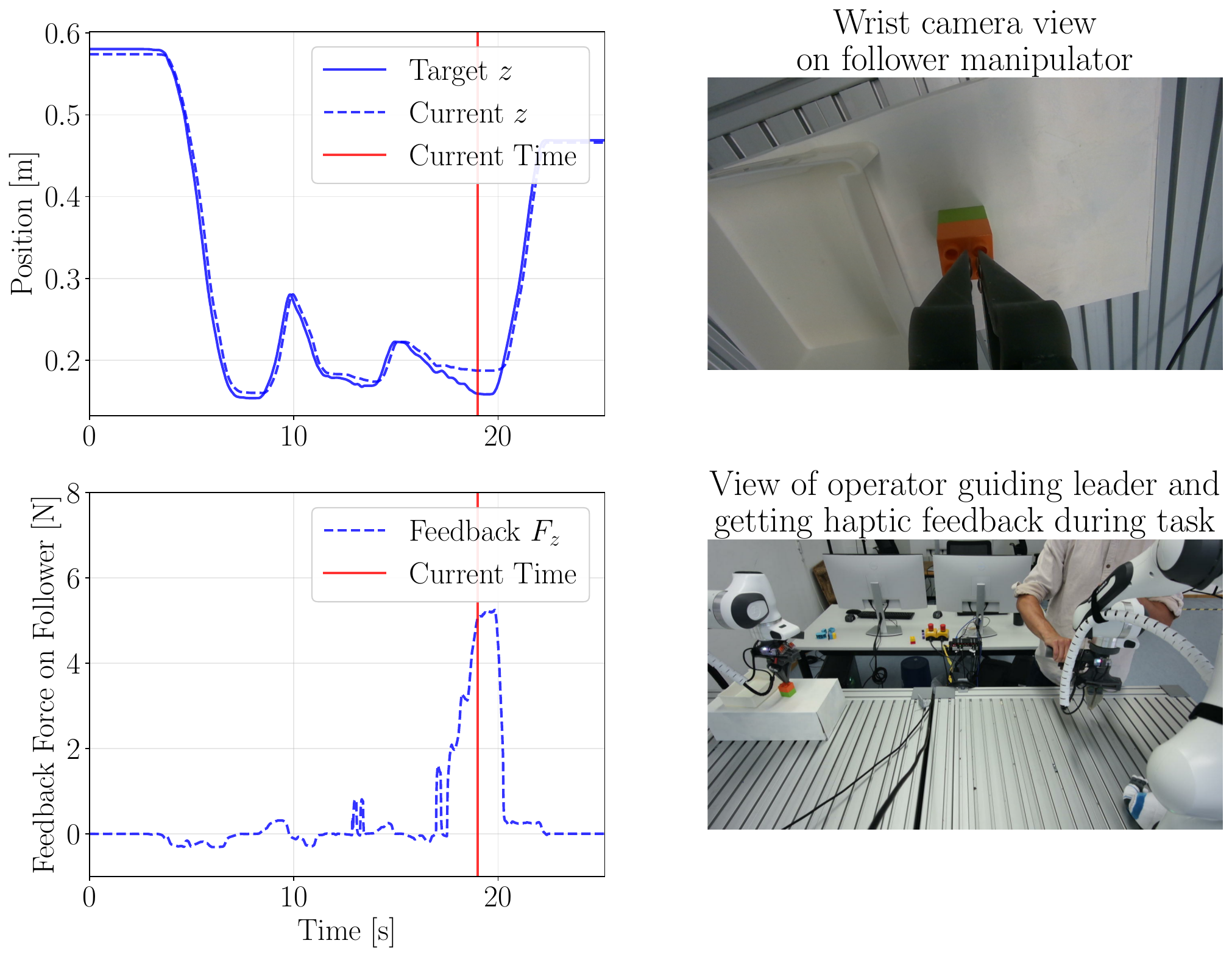}
    \caption{\textbf{Unified data collection and policy deployment [I]:} Recording a block-stacking task with Lego using teleoperation and force-torque feedback for the operator.  
    \textbf{Upper left:} Plot of the follower’s target and end-effector positions.  
    \textbf{Lower left:} Feedback wrenches applied at the leader manipulator.  
    \textbf{Upper right:} Image of the recording on the follower manipulator at the current time.  
    \textbf{Lower right:} Image of the operator guiding the leader robot to record data on the follower at the current time.}
    \label{fig:teleoperation}
    \vspace{-2pt}
\end{figure}

Second, we demonstrate our teleoperation setup for a Lego stacking task in ~\autoref{fig:teleoperation}. During teleoperation, the end effector poses from the leader robot are streamed to the follower at roughly 30 Hz, where they are tracked using the \textsc{CRISP} CI controller. The torque-based controllers allow for force feedback as well as compliant behavior.

\begin{figure}[t]
    \centering
    \includegraphics[width=1\linewidth]{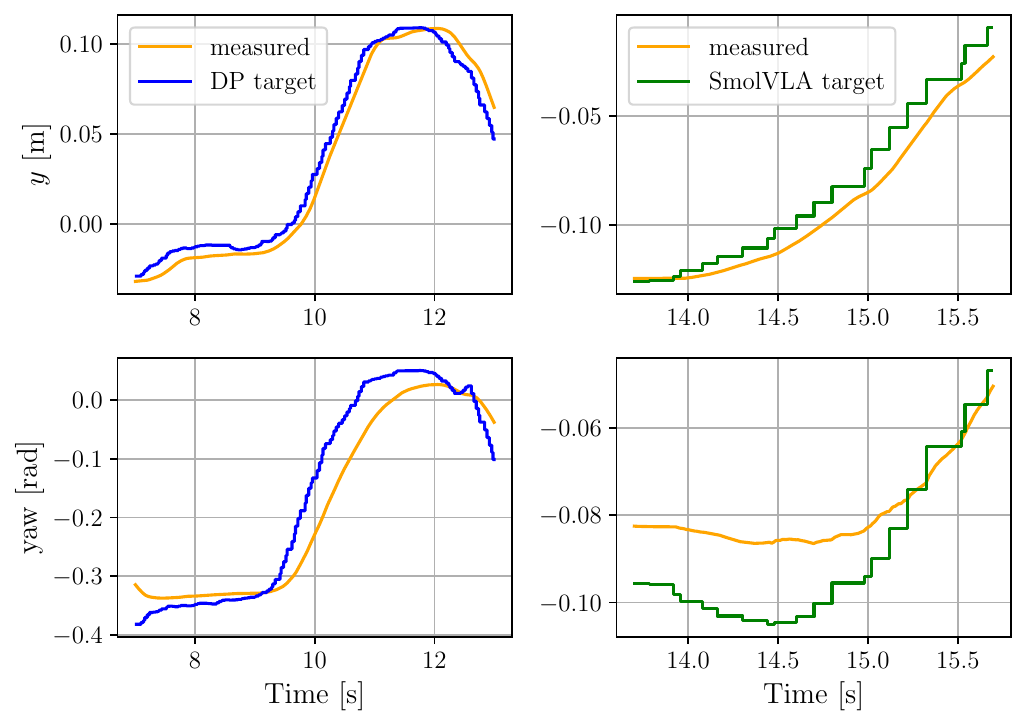}
    \caption{\textbf{Unified data collection and policy deployment [II]:}  
    Using the same recorded data in the \textsc{CRISP\_GYM} interface, different learned policies can be successfully deployed in hardware.
    \textbf{Left:} Diffusion Policy \cite{chi2024diffusionpolicy} operating at approximately 30~Hz.  
    \textbf{Right:} SmolVLA \cite{shukor2025smolvla} operating at approximately 10~Hz.  
    Both policies provide end-effector target pose updates that are tracked by our \textsc{CRISP} controllers. Note the different timescales shown in the plots.  
    For policy deployment, we used the CI controllers with \textsc{CRISP\_GYM}.}

    \label{fig:policies}
    \vspace{-12pt}
\end{figure}

Third, we evaluate our controllers deployed with two imitation learning policies, namely Diffusion Policy~\cite{chi2024diffusionpolicy} and SmolVLA~\cite{shukor2025smolvla}, running at different inference frequencies on a GPU. The policies are deployed with the same Python interface as used for the previously described data collection with teleoperation. 
The results are shown in~\autoref{fig:policies}.

Our experiments demonstrate that the proposed controllers are suitable to track poses commanded by learning-based policies or by a human operator through teleoperation.

\section{Conclusion}

In this work, we present CRISP - an implementation of controllers for robotic manipulators designed for the deployment of learning-based policies that require tracking of small action chunks or low-frequency, discontinuous target joints or poses. The robot-agnostic implementation in ROS2 enables broad applicability across platforms, while its lightweight software design facilitates easy integration in new software stacks. Our framework can be used for both data collection and policy deployment across different manipulators, thereby accelerating research in learning-based manipulation.

\bibliographystyle{IEEEtran}
\bibliography{references} 

\end{document}